\documentclass{spie}

\usepackage{cite}

\usepackage{times} 
\usepackage{indentfirst} 

\usepackage{todonotes}

\presetkeys%
    {todonotes}%
    {inline}{}
    
\usepackage{mathtools}

\usepackage{enumitem}

\usepackage[hyphens]{url}
\usepackage[hidelinks]{hyperref}
\usepackage{import}

\usepackage{hyperref}

\hypersetup{breaklinks=true}
\usepackage{multirow}

\DeclarePairedDelimiterX{\norm}[1]{\lVert}{\rVert}{#1}

\title{Multiple Light Source Dataset for Colour Research}
\author{Anna Smagina\supit{1}, Egor Ershov\supit{1}, Anton Grigoryev\supit{1, 2}
    \skiplinehalf
    \normalsize 
    \supit{1}Institute for Information Transmission Problems (Kharkevich Institute) of the Russian Academy of Sciences \\
    \supit{2}Moscow Institute of Physics and Technology (State University)
  }

\begin{document}
    \maketitle
 

    \begin{abstract}
We present a collection of 24 multiple object scenes recorded under 18 multiple light source illumination scenarios each.
The illuminants are varying in dominant spectral colours, intensity and distance from the scene.
We mainly address the realistic scenarios for evaluation of computational colour constancy algorithms, but also have aimed to make the data as general as possible for computational colour science and computer vision.
Along with the images, we provide also spectral characteristics of the camera, light sources, and the objects and include pixel-by-pixel ground truth annotation of uniformly coloured object surfaces.
The dataset is freely available at \url{https://github.com/visillect/mls-dataset}.
\end{abstract}
    \section{Introduction}

In this paper we describe a new laboratory dataset mainly designed for evaluation of computational colour constancy\cite{gijsenij2011computational, nikolayev2007new}. 
39 years have already passed since the GrayWorld method invention \cite{buchsbaum1980spatial}, which is one of the first and the most popular colour constancy approaches.
Colour constancy algorithms find their application in such tasks as face identification\cite{sengupta2018sfsnet}, illumination-invariant object tracking \cite{agarwal2006overview}, shadow suppression for image segmentation\cite{zhou2015moving}.
The research activity in this topic is not declining even now, which is largely due to the increase in available computing power and the to development of technologies for creating better optical sensors.
One can learn more about the history of the  colour constancy research up to $2014$ from the work \cite{gijsenij2011computational}.

Indicative of the relevance of colour constancy to modern science is the fact that in the period from $2002$ to the present, on average, each year one colour constancy dataset was published.
Such datasets could be essentially be divided into two types: ones collected in the laboratory and ones collected under uncontrolled conditions.
And while the latter are usually created to evaluate existing algorithms under real-life conditions, the former are more likely to accurately evaluate the quality of existing solutions, as well as to determine solvability of tasks that have not yet been solved.
The list of the latter should include such problems as the colour estimation of multiple light sources (about ten works have already been written on this topic, in particular \cite{bianco2017single}), restoring the position of light sources in the camera coordinate system from an image, creating low-parametric spectral models for describing colour transformations \cite{gusamutdinova2017verification}, and others.

Datasets collected under natural and uncontrolled conditions, for example, REC \cite{hemrit2018rehabilitating}, CubePlus \cite{banic2017unsupervised}, GrayBall \cite{ciurea2003large}, \mbox{NUS \cite{cheng2014illuminant}}, are poorly suited for quality evaluation of problems listed above.
The estimation of source colour in such datasets is performed using the calibration object (or objects) illuminated by several light sources simultaneously. 
But since its proportions are unknown, the estimation of the source colour is performed inaccurately.
The same is true for sets of optical multichannel \mbox{images \cite{nascimento2002statistics, foster2006frequency, parraga1998color}}.
Although they allow not only a more accurate colorimetric algorithms evaluation, but also an augmentation of lighting.
For example, \cite{gijsenij2011color} proposes a method for synthesis images such that different parts of the scene are illuminated by various sources.

More precisely, the illuminant estimation can be performed either via increasing the number of calibration objects in the scene, for example, this is done in \cite{funt2010rehabilitation}, or, more accurately but more time-consuming, by measuring the emission spectra of illuminants (our method).
While forming this dataset, we were focusing precisely on accurate measuring of all characteristics in the scene, both spectra and the location of objects.
In the context of the tasks mentioned above, the following requirements were defined for the dataset:

\begin{itemize}[noitemsep]
    \item The presence of different light sources (different chromaticities) with known spectral characteristics;
    \item The presence of scenes illuminated by one or several light sources of various types (spot and diffuse);
    \item Various object shapes and materials---metals (preferably chromatic), dielectrics, flat / non-flat (important for evaluation of algorithms based on reflectance models \cite{finlayson2001solving, brill1990image});
    \item Known spectral characteristics of the objects reflectance.
    \item Known camera characteristics and linear images.
\end{itemize}
Among the published laboratory datasets \cite{barnard2002comparison, geusebroek2005amsterdam, bleier2011color, rizzi2013yaccd2} there is no one that would simultaneously satisfy all the items on the list, which served as the motivation for the construction of the one described in this paper.
Although, collected data is also usefull for colour-based image segmentation, validation of spectral colour models, and other colour vision problems.

    \section{Scene setup}
\label{scene_setup}

\begin{figure}[!h]
    \centering
    \includegraphics[width=0.85\linewidth]{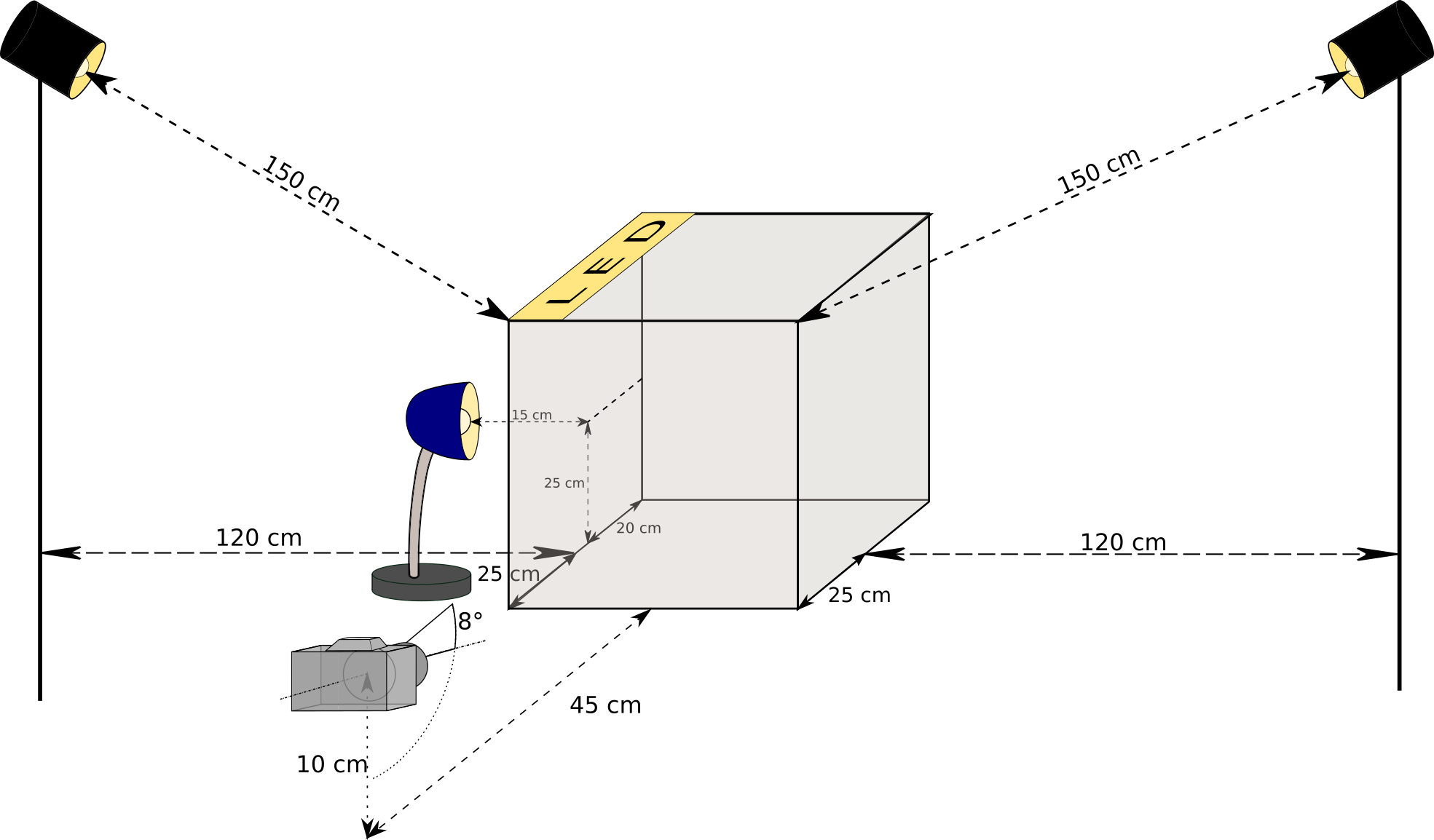} \\
    \caption{\label{fig:dataset_experimental_setup}
 Schematic view of the experimental setup for image collection.}
\end{figure}

The setup used to acquire the image collection consists of a softbox (FALCON EYES PBF-60AB), 4 different light sources and a camera, see Fig. \ref{fig:dataset_experimental_setup}.
The halogen lights are mounted at equal distances (1.5 m) from two opposite sides of the softbox to produce a diffuse lighting.
In immediate vicinity of the softbox the desk and the LED lamps are placed to produce close lighting.
Objects are positioned manually inside the of the softbox. 
The camera is pointing on the scene through the open side of the softbox.

The halogen lamps are 35 W each with similar light spectra. 
Duration of the whole experimental procedure (setting-up, calibrating, and recording the database, approx. 100 hours) was only a fraction of the average life time of the lamps (about 1000 hours), hence ageing effects affecting colour temperature can be considered minimal. 
The brightness of the incandescent lamp (60W) was adjustable and was set to 30\% of the maximum.
The RGB LED strip is under computer control via in-house made USB controller, which allows to tune the emitting power of red, green and blue LEDs individually.

Images are recorded with Canon 5D Mark III camera equipped with Canon EF 24–70 mm f/2.8L II USM lens.
\mbox{The shooting} was performed in a RAW format with 5760 x 3840 resolution in the manual mode. 
The camera settings was adjusted under the maximal bright in the experiment illumination in the way that avoids over-exposures.
Zoom was switched off. 
Focal length was set to 25 mm, ISO speed to 320. Aperture was closed to F/16 to achieve sharpness across most of the scene.
The exposure time was set to 1 second and the 2 second delay was used for shooting.
The in-camera white balance was adjusted to 6500K.
The spectral responses for Canon 5D Mark III was measured by Baek et al. in \cite{baek2017compact}. 
To extract this data published as figures we use Web Plot Digitizer \cite{webplotdigitizer}.
The resulting csv-files are provided along with the dataset.

    \section{Objects}

To compose the scenes we used objects without colour-textured surfaces and mirroring effects to provide clear colouring annotation (section \ref{image_data}) and spectral characteristics (section \ref{spectral_data}).
The objects represent 122 surfaces of various materials and shapes.
These include 88 dielectric surfaces (both matte and glance), 6 achromatic and 10 chromatic metals.
Also, we also recorded a scene with DGK Color Tools WDKK Color Chart consisting of 18 patches with known colors (fig. \ref{fig:lighting_overview}).
    \section{Image data}
\label{image_data}

\begin{figure}[b]
    \centering
    
    \minipage{0.28\linewidth}
        \vspace{-17ex}
        \minipage{\linewidth} \center{\includegraphics[width=0.98\linewidth]{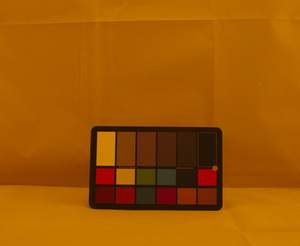}}    \endminipage 
        \vspace{0.2ex} 
        \\
        \minipage{\linewidth} \center{\scriptsize{2HAL}} \endminipage 
        \vspace{0.8ex} 
        \\
        \minipage{\linewidth} \center{\includegraphics[width=0.98\linewidth]{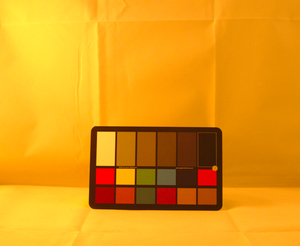}} \endminipage   
        \vspace{0.2ex} 
        \\
        \minipage{\linewidth} \center{\scriptsize{2HAL\_DESK}} \endminipage
    \endminipage
    \hfill
    \minipage{0.7\linewidth} 
        \minipage{0.24\linewidth} \center{\includegraphics[width=0.98\linewidth]{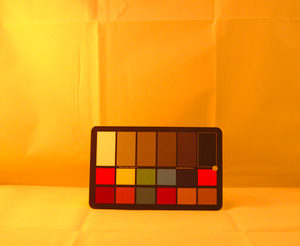}} \endminipage 
        \minipage{0.24\linewidth} \center{\includegraphics[width=0.98\linewidth]{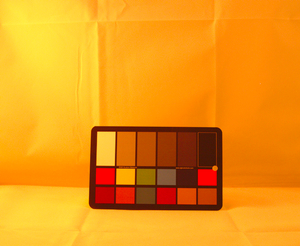}} \endminipage 
        \minipage{0.24\linewidth} \center{\includegraphics[width=0.98\linewidth]{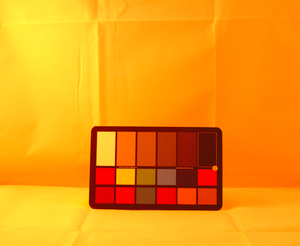}} \endminipage 
        \minipage{0.24\linewidth} \center{\includegraphics[width=0.98\linewidth]{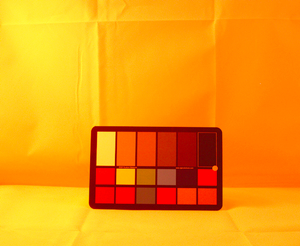}} \endminipage     
        \vspace{0.2ex} 
        \\
        \minipage{0.24\linewidth} \center{\scriptsize{2HAL\_DESK\_LED-R025}} \endminipage 
        \minipage{0.24\linewidth} \center{\scriptsize{2HAL\_DESK\_LED-R050}} \endminipage     
        \minipage{0.24\linewidth} \center{\scriptsize{2HAL\_DESK\_LED-R075}} \endminipage 
        \minipage{0.24\linewidth} \center{\scriptsize{2HAL\_DESK\_LED-R100}} \endminipage               
        \vspace{0.6ex} 
        \\
        \minipage{0.24\linewidth} \center{\includegraphics[width=0.98\linewidth]{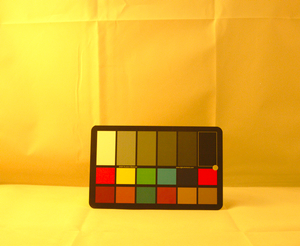}} \endminipage 
        \minipage{0.24\linewidth} \center{\includegraphics[width=0.98\linewidth]{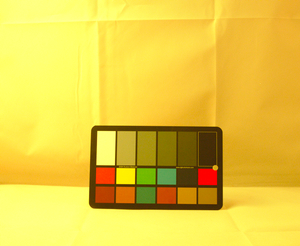}} \endminipage 
        \minipage{0.24\linewidth} \center{\includegraphics[width=0.98\linewidth]{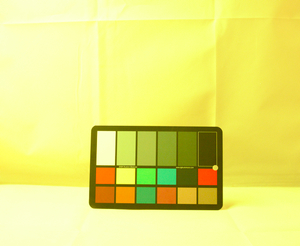}} \endminipage 
        \minipage{0.24\linewidth} \center{\includegraphics[width=0.98\linewidth]{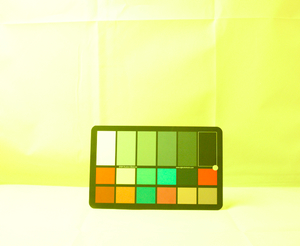}} \endminipage    
        \vspace{0.2ex} 
        \\
        \minipage{0.24\linewidth} \center{\scriptsize{2HAL\_DESK\_LED-RG025}} \endminipage 
        \minipage{0.24\linewidth} \center{\scriptsize{2HAL\_DESK\_LED-RG050}} \endminipage     
        \minipage{0.24\linewidth} \center{\scriptsize{2HAL\_DESK\_LED-RG075}} \endminipage 
        \minipage{0.24\linewidth} \center{\scriptsize{2HAL\_DESK\_LED-RG100}} \endminipage               
        \vspace{0.6ex} 
        \\
        \minipage{0.24\linewidth} \center{\includegraphics[width=0.98\linewidth]{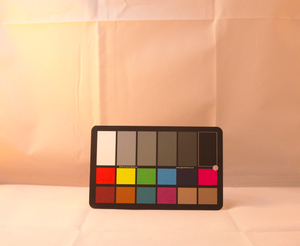}} \endminipage 
        \minipage{0.24\linewidth} \center{\includegraphics[width=0.98\linewidth]{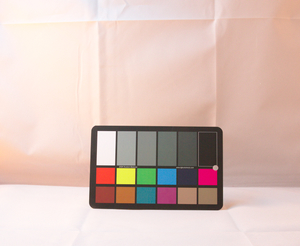}} \endminipage 
        \minipage{0.24\linewidth} \center{\includegraphics[width=0.98\linewidth]{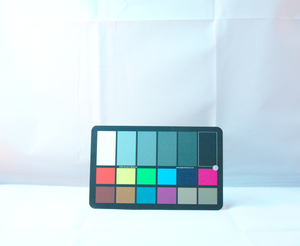}} \endminipage 
        \minipage{0.24\linewidth} \center{\includegraphics[width=0.98\linewidth]{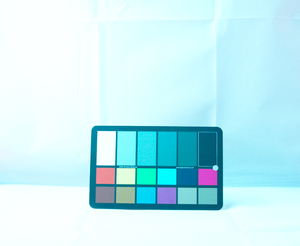}} \endminipage 
        \vspace{0.2ex} 
        \\
        \minipage{0.24\linewidth} \center{\scriptsize{2HAL\_DESK\_LED-BG025}} \endminipage 
        \minipage{0.24\linewidth} \center{\scriptsize{2HAL\_DESK\_LED-BG050}} \endminipage     
        \minipage{0.24\linewidth} \center{\scriptsize{2HAL\_DESK\_LED-BG075}} \endminipage 
        \minipage{0.24\linewidth} \center{\scriptsize{2HAL\_DESK\_LED-BG100}} \endminipage       
        \vspace{0.6ex} 
        \\
        \minipage{0.24\linewidth} \center{\includegraphics[width=0.98\linewidth]{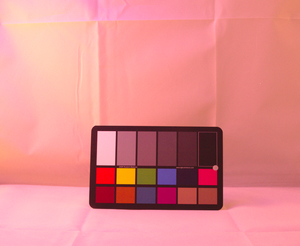}} \endminipage 
        \minipage{0.24\linewidth} \center{\includegraphics[width=0.98\linewidth]{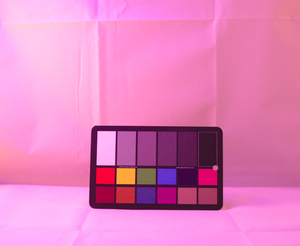}} \endminipage 
        \minipage{0.24\linewidth} \center{\includegraphics[width=0.98\linewidth]{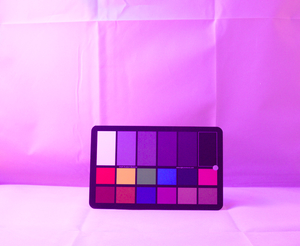}} \endminipage 
        \minipage{0.24\linewidth} \center{\includegraphics[width=0.98\linewidth]{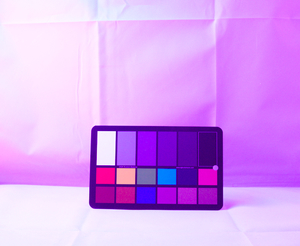}} \endminipage 
        \vspace{0.2ex} 
        \\
        \minipage{0.24\linewidth} \center{\scriptsize{2HAL\_DESK\_LED-B025}} \endminipage 
        \minipage{0.24\linewidth} \center{\scriptsize{2HAL\_DESK\_LED-B050}} \endminipage     
        \minipage{0.24\linewidth} \center{\scriptsize{2HAL\_DESK\_LED-B075}} \endminipage 
        \minipage{0.24\linewidth} \center{\scriptsize{2HAL\_DESK\_LED-B100}} \endminipage       
        \\
    \endminipage     

    \vspace{0.4ex}          
    \caption{    \label{fig:lighting_overview}
 Example scene viewed under 18 different illumination scenarios.}
\end{figure}

The scenes were composed from multiple objects placed on a coloured background.
The illumination of each scene was varied in 18 configurations, which are shown at figure \ref{fig:lighting_overview} and include the following combinations of light sources used for shooting:

\begin{itemize}[noitemsep, nolistsep]
    \item \texttt{2HAL} --- two halogen lights only, 
    \item \texttt{2HAL\_DESK} -- two halogen lights with the desktop lamp,
    \item \texttt{2HAL\_DESK\_LED-\{R, RG, BG, B\}\{025, 050, 075, 100\}} --- two halogen lights with the desktop lamp and red (R), blue (B), red and green (RG) or blue and green (BG) lights of the LED strip turned on at 25\%, 50\%, 70\%, and 100\% of the emitting power correspondingly.
\end{itemize}


Each image is provided in 3 formats: a full-resolution raw (single-channel RGGB Bayer array) image cropped and converted to 16-bit PNG format from the camera-specific raw CR2 image, a half-resolution colour PNG (in linear sensor-specific RGB coordinates) image obtained by naive demosaicing algorithm (the camera sensor has a built-in optical low-pass filter, so no moire artifacts), and a quarter-resolution colour JPEG image for preview. The raw and 16-bit colour images are presented without any auto-contrast or range stretching and may appear very dark when viewed in ordinary software, however they contain all information to reconstruct the full-colour images as given by previews.

Each scene is accompanied with the pixel-wise annotation masks of uniformly coloured object surfaces.
The masks are given in a 8-bit PNG file, where pixels corresponding to uniformly coloured object surface have the same, unique on the mask, colour. 
For annotation purposes, images were first automatically split into small regions (but not less than 100 pixels) with guaranteed colour constancy using the algorithm given in \cite{smagina2019linear}.
Then, regions were manually merged in accordance to uniform colouring of the scene.
Each such region is associated with a reflectance spectrum (section \ref{spectral_data}).
Same as with the scene images, the colouring annotation is provided in full-, half- and quarter-resolutions.


\begin{figure}[t]
   
    \centering
    \minipage{0.2\linewidth} \center{\includegraphics[width=0.98\linewidth]{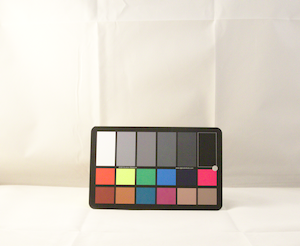}} \endminipage     
    \minipage{0.2\linewidth} \center{\includegraphics[width=0.98\linewidth]{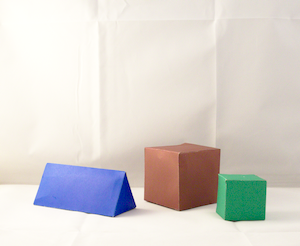}} \endminipage 
    \minipage{0.2\linewidth} \center{\includegraphics[width=0.98\linewidth]{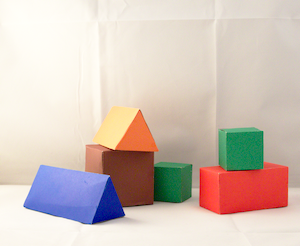}} \endminipage 
    \minipage{0.2\linewidth} \center{\includegraphics[width=0.98\linewidth]{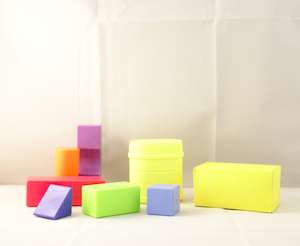}} \endminipage 
    \\
    \vspace{0.1ex}      
    \minipage{0.2\linewidth} \center{\includegraphics[width=0.98\linewidth]{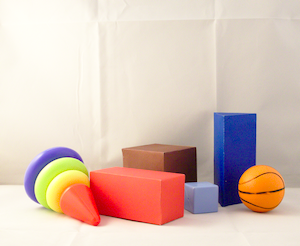}} \endminipage     
    \minipage{0.2\linewidth} \center{\includegraphics[width=0.98\linewidth]{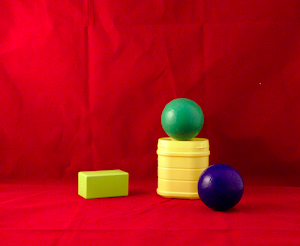}} \endminipage 
    \minipage{0.2\linewidth} \center{\includegraphics[width=0.98\linewidth]{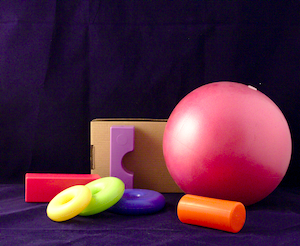}} \endminipage 
    \minipage{0.2\linewidth} \center{\includegraphics[width=0.98\linewidth]{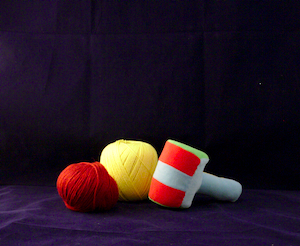}} \endminipage 
    \\
    \vspace{0.1ex}
    \minipage{0.2\linewidth} \center{\includegraphics[width=0.98\linewidth]{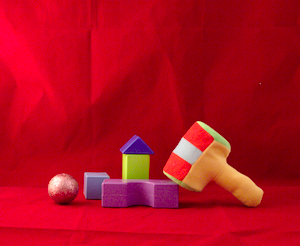}} \endminipage     
    \minipage{0.2\linewidth} \center{\includegraphics[width=0.98\linewidth]{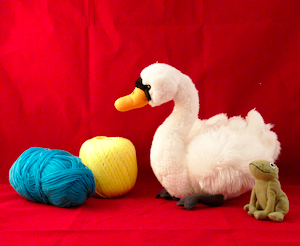}} \endminipage 
    \minipage{0.2\linewidth} \center{\includegraphics[width=0.98\linewidth]{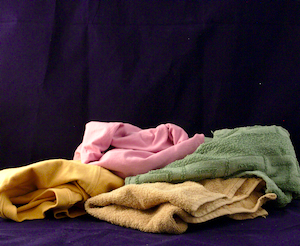}} \endminipage 
    \minipage{0.2\linewidth} \center{\includegraphics[width=0.98\linewidth]{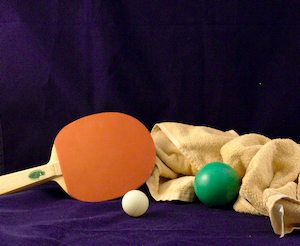}} \endminipage 
    \\
    \vspace{0.1ex}
    \minipage{0.2\linewidth} \center{\includegraphics[width=0.98\linewidth]{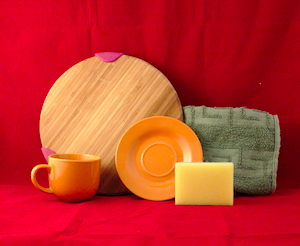}} \endminipage     
    \minipage{0.2\linewidth} \center{\includegraphics[width=0.98\linewidth]{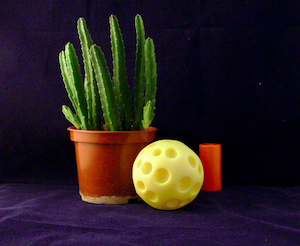}} \endminipage 
    \minipage{0.2\linewidth} \center{\includegraphics[width=0.98\linewidth]{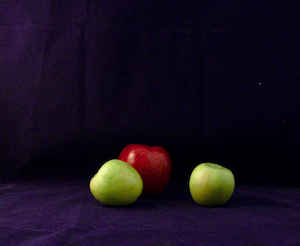}} \endminipage 
    \minipage{0.2\linewidth} \center{\includegraphics[width=0.98\linewidth]{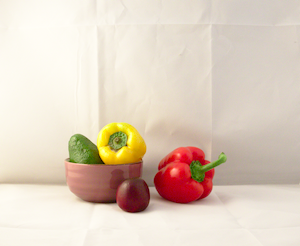}} \endminipage 
    \\
    \vspace{0.1ex}    
    \minipage{0.2\linewidth} \center{\includegraphics[width=0.98\linewidth]{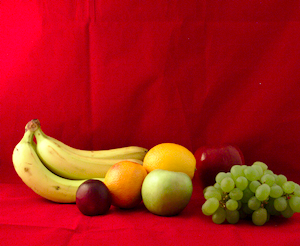}} \endminipage     
    \minipage{0.2\linewidth} \center{\includegraphics[width=0.98\linewidth]{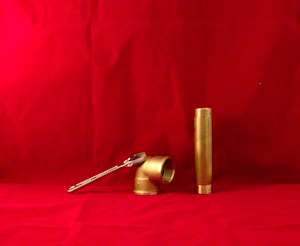}} \endminipage 
    \minipage{0.2\linewidth} \center{\includegraphics[width=0.98\linewidth]{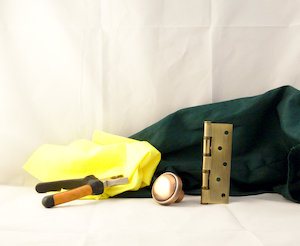}} \endminipage 
    \minipage{0.2\linewidth} \center{\includegraphics[width=0.98\linewidth]{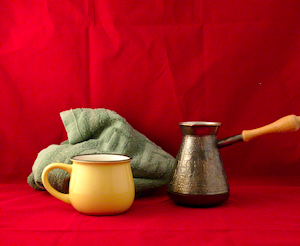}} \endminipage 
    \\
    \minipage{0.2\linewidth} \center{\includegraphics[width=0.98\linewidth]{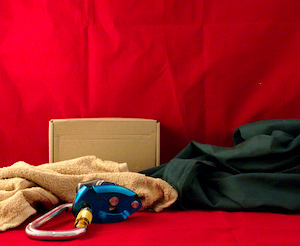}} \endminipage     
    \minipage{0.2\linewidth} \center{\includegraphics[width=0.98\linewidth]{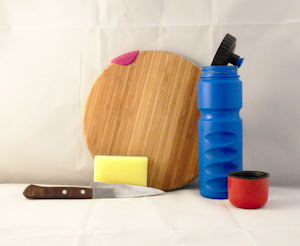}} \endminipage 
    \minipage{0.2\linewidth} \center{\includegraphics[width=0.98\linewidth]{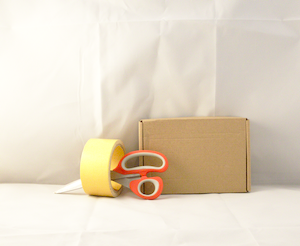}} \endminipage 
    \minipage{0.2\linewidth} \center{\includegraphics[width=0.98\linewidth]{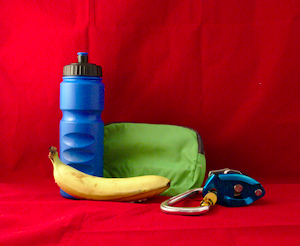}} \endminipage 
    \\
    \vspace{1ex}
    \caption{ \label{fig:scenes_overview} Overview of all recorded MLS scenes.
    Here the white balancing is performed for demonstration purpose.}
\end{figure}
    \section{Spectral data} 
\label{spectral_data}

In addition to colour images under various lighting conditions, we also collected a set of spectral measurements, both for illuminant emission and for object reflectance. The measurement was conducted using a miniature spectrometer OceanOptics FLAME-S-VIS-NIR-ES with a corresponding set of accessories from OceanOptics, including two light sources (HL-2000 for reflectance measurement and HL-3P-CAM for radiometric calibration for illuminant characterization) and a reflectance standard WS-1.

The source spectra were collected for all individual sources (two halogen lamps, the desk lamp and the 3 components of the RGB LED lamp) measured through the softbox, to characterise the real spectra illuminating the objects. Also, the spectrum of both halogen lamps measured near the camera lens to get some information on relative spectral intensities. The dataset is accompanied by the photos illustrating the spectral measurement conditions.

\begin{figure}[h!]
    \center
    \minipage{0.44\linewidth} \center{Chromaticities of illuminants} \endminipage 
    \minipage{0.44\linewidth} \center{Chromaticities of surfaces} \endminipage         
    \\
    \minipage{0.44\linewidth} \center{\includegraphics[width=0.98\linewidth]{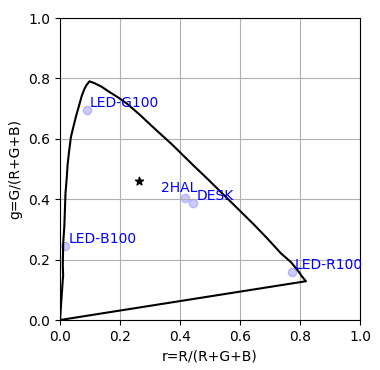}} \endminipage 
    \minipage{0.44\linewidth} \center{\includegraphics[width=0.98\linewidth]{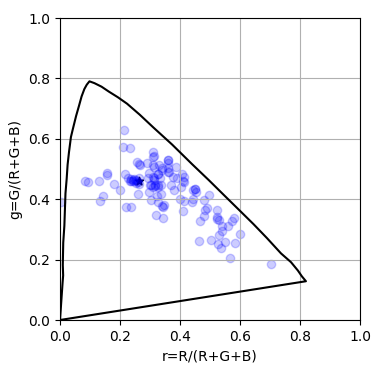}} \endminipage 
    \\
    \minipage{0.44\linewidth} \center{(a)} \endminipage 
    \minipage{0.44\linewidth} \center{(b)} \endminipage         
    \\    
    \vspace{0.6ex}          
    \caption{    \label{fig:spectra_chromacities}
 Chromaticity distributions of illuminants and surfaces in camera-specific colour coordinates. 
    The star marker denotes the white point. }    
\end{figure}

The measurements were taken with built-in non-linearity correction and calibrated for the uniform sensitivity for all wavelengths using the radiometrically calibrated light source. The dark room background was taken into account and subtracted, however, some small non-uniformity (e.g. imperfectly blinking indicator lights on) can be visible in the resulting spectra.
The chromaticities of illuminants relative to our camera are shown in the figure \ref{fig:spectra_chromacities}a.

The reflectance spectra were measured relative to the polytetrafluoroethylene standard (Ocean Optics WS-1). The reflection probe holder was used to control the angle of the measurement. 
$45^{\circ}$ measurement angle was used for all materials, the specular component of metallic or metallised surfaces was also measured with probe perpendicular to the surface.
For each object the spectra for all major constituent materials visible in the scene were measured.
Chromacities of all surfaces taken at $45^{\circ}$ measurement angle are shown in the figure \ref{fig:spectra_chromacities}b.
The materials are specified by a unique string (e.g. ``red\_plastic\_ball'' or ``brown\_metallic\_doorknob'') containing the colour, the material and the object identification.

    \section{Consistency of spectral and image data}

Here we demonstrate the expected accuracy in illuminant or surface chromaticity estimation using this data, to provide a better support for benchmarking of colour vision algorithms.

\begin{figure}[h!]
    \center
    \minipage{0.44\linewidth}\center{No chomaticity mapping}\endminipage
    \minipage{0.44\linewidth}\center{Fitting with colour homography}\endminipage     
    \\  
    \minipage{0.44\linewidth} 
    \center{\includegraphics[width=0.98\linewidth]{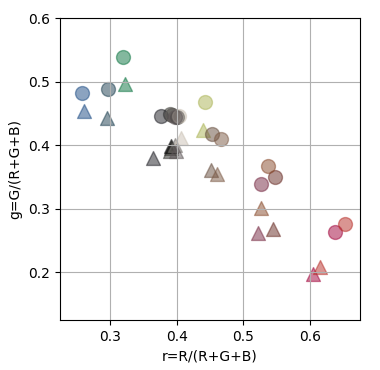}} \endminipage 
    \minipage{0.44\linewidth} 
    \center{\includegraphics[width=0.98\linewidth]{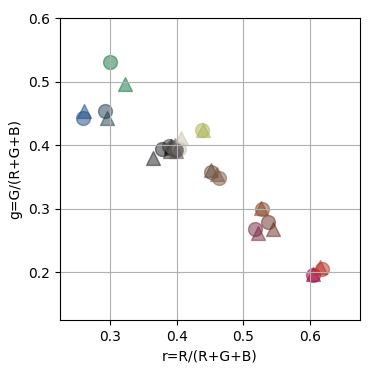}} 
    \endminipage 
    \\
    \minipage{0.44\linewidth}\center{(a)}\endminipage
    \minipage{0.44\linewidth}\center{(b)}\endminipage     
    \\    
    \caption{\label{fig:consisitency_check}
    Chromaticities of colour chart patches in camera-specific colour coordinates calculated from measured spectra (marked with triangles) and captured images (marked with circles).
    The marker colours correspond to colours of the patches.
    On the plot (a) no transform is applied to patches chromaticities coordinates both for measured and observed in the image.
    On the plot (b) we applied homography to observed patches chromaticities coordinates.}
\end{figure}

For demonstration we used the image of the colour chart taken under \texttt{2HAL} illumination (Fig. \ref{fig:lighting_overview}). 
For each patch we calculated the chromaticity in a two ways:
first, using \textit{measured} spectral data (surface spectrum, illuminant spectrum and camera spectral responses), second, directly from the RGB image -- \textit{observed} chromaticities.
The former was obtained in terms of Lambertian image formation model.
The result is given in the figure \ref{fig:consisitency_check}a.
As it could be seen, the chromaticity coordinates of white and neutral patches are densely grouped together, while the coordinates of light blue (on the left) and red, brown, and purple patches (on the ride side of the chromaticity distribution) matches poorly.

Miscalculations, shown in figure \ref{fig:consisitency_check}a, could be at least partially explained by linear miscalculations in camera spectral sensitivity and lighting setup. 
Considering, that the chomaticities across a linear change in capture condition (light color, shading and imaging device) are homography apart \cite{finlayson2017color}, 
we applied better fitting of observed chromaticity coordinates to the measured ones. 
As it shown in figure \ref{fig:consisitency_check}b, the homography transform of observed chromaticitiy coordinates provides visibly better correspondence -- the measured and observed coordinates on the left and the right side of the chromaticity distribution now more closely located to each other.
To estimate planar homography matrix we conduct fitting above all of the patches chromaticity coordinates. 

However, some inaccuracies are still remain, which are probably caused by non-linear effects in image noise -- such as, for example, because of the colour chart was glance, not matte.
Also note, that in this work we did not measured the camera responses by ourselves (section 2).

    \section{Conclusion}

We present a collection of 24 scenes images recorded under 18 different multiple light source illuminations. 
The main feature of the dataset is a completeness, containing spectral data both for illuminants and for object surfaces.
The dataset mainly offers a evaluation for computation colour constancy, but also can be used in variety of computer vision involving a colour- or material based image segmentation, photometric invariant extraction and others.
    \section*{ACKNOWLEDGEMENTS}

This work is partially supported by Russian Foundation for Basic Research (projects 17-29-03514 and 17-29-03370).

We are grateful to Alexander Belokopytov for his help in constructing the acquisition setup. 
We thank Dmitry Nikolaev for his valuable comments in experimental design and result verification.
We thank Alexey Glikin for the technical help. 
For contributing the objects, we acknowledge Artem Sereda, Sergey Gladilin, Dmitry Bocharov.

    \bibliographystyle{spiebib}
    \bibliography{bibliography}

    
\end{document}